\documentclass{article}

\usepackage{authblk}

\usepackage{arxiv}

\usepackage[utf8]{inputenc} 
\usepackage[T1]{fontenc}    
\usepackage{hyperref}       
\usepackage{url}            
\usepackage{booktabs}       
\usepackage{amsfonts}       
\usepackage{amssymb}        
\usepackage{nicefrac}       
\usepackage{microtype}      
\usepackage{graphicx}
\usepackage{doi}
\usepackage{xcolor}
\usepackage{longtable}   
\usepackage{booktabs}    
\usepackage{pdflscape}   
\usepackage{caption}
\usepackage{subcaption}
\usepackage{enumitem}       

\newenvironment{compactitem}
{\scriptsize\begin{itemize}[nosep, leftmargin=1em, labelsep=0.5em, topsep=0pt, partopsep=0pt]}
{\end{itemize}}

\usepackage[final]{changes}
\usepackage{lineno}
\definechangesauthor[name={slimane}, color=red]{sbm}
\definechangesauthor[color=green]{bb} 
\definechangesauthor[color=blue]{ihb} 

\title{A Reproducible Framework for Neural Topic Modeling in Focus Group Analysis}

\author[1,2]{Heger Arfaoui}
\author[3]{Mohamed Iheb Hergli}
\author[2]{Beya Benzina}
\author[2]{Slimane BenMiled}

\affil[1]{National Engineering School of Tunis, University of Tunis El Manar, Tunis, Tunisia}
\affil[2]{Pasteur Institute of Tunis, University of Tunis El Manar, Tunis, Tunisia}
\affil[3]{Mediterranean Institute of Technology, South Mediterranean University, Tunis, Tunisia}

\date{}

\hypersetup{
pdftitle={A Reproducible Framework for Neural Topic Modeling in Focus Group Analysis},
pdfsubject={},
pdfauthor={Heger Arfaoui, Mohammed Iheb Hergli, Beya Benzina, Slimane BenMiled},
}

\begin{document}
\maketitle

\begin{abstract}
Focus group discussions generate rich qualitative data but their analysis traditionally relies on labor-intensive manual coding that limits scalability and reproducibility. We present a systematic framework for applying BERTopic to focus group transcripts using data from ten focus groups exploring HPV vaccine perceptions in Tunisia (1,075 utterances). We conducted comprehensive hyperparameter exploration across 27 configurations, evaluating each through bootstrap stability analysis, performance metrics, and comparison with LDA baseline. Bootstrap analysis revealed that stability metrics (NMI and ARI) exhibited strong disagreement (r = -0.691) and showed divergent relationships with coherence, demonstrating that stability is multifaceted rather than monolithic. Our multi-criteria selection framework yielded a 7-topic model achieving 18\% higher coherence than optimized LDA (0.573 vs. 0.486) with interpretable topics validated through independent human evaluation (ICC = 0.700, weighted Cohen's $\kappa$ = 0.678). These findings demonstrate that transformer-based topic modeling can extract interpretable themes from small focus group transcript corpora when systematically configured and validated, while revealing that quality metrics capture distinct, sometimes conflicting constructs requiring multi-criteria evaluation. We provide complete documentation and code to support reproducibility.

\end{abstract}


\section{Introduction}
\label{sec:introduction}

Topic modeling has emerged as a powerful computational approach for discovering latent thematic structures in large text corpora. Originally developed for document organization and information retrieval~\cite{blei2003}, topic modeling techniques have found widespread application across diverse text types. Early work focused on structured document collections such as news articles, scientific papers, and web pages~\cite{griffiths2004, newman2006}. More recently, researchers have applied topic modeling extensively to social media platforms including Twitter~\cite{qorib_covid-19_2023, lyu_covid-19_2021}, Reddit~\cite{fong2025, yang2025}, and online forums~\cite{sik2023, davila2023}, as well as to customer reviews~\cite{yazici2024, liu2023} and open-ended survey responses~\cite{guetterman_augmenting_2018}. Recent advances in transformer-based architectures have enabled neural topic models that leverage contextual embeddings rather than traditional bag-of-words representations, offering improved performance for short, noisy text common in social media~\cite{grootendorst2022, egger2022}.

Despite this breadth of applications, most implementations target large-scale corpora where statistical inference benefits from thousands or tens of thousands of documents. Moreover, standard benchmarks and evaluation protocols emphasize either well-structured documents (e.g., 20 Newsgroups, scientific abstracts) or homogeneous short texts (e.g., tweets about specific events). This leaves open questions about topic modeling performance on small, heterogeneous conversational corpora where documents vary substantially in length and thematic focus, and where corpus size may comprise only hundreds of utterances.

Focus group discussions represent a particularly important yet computationally underexplored text type. Unlike social media posts, which are spontaneous and often brief, focus groups are structured conversations where participants engage in moderated dialogue following a semi-structured protocol. Unlike customer reviews or survey responses, which target specific products or questions, focus groups explore open-ended topics through emergent conversation. Focus groups are widely used across health research, market research, policy analysis, and user experience studies, generating rich textual data that captures nuanced perspectives through natural conversation. Despite their methodological prevalence, focus group transcripts have received limited attention in the computational topic modeling literature, likely reflecting both the proprietary nature of such data and the traditional emphasis on manual thematic coding. Yet when research objectives include systematic synthesis across multiple sessions, geographic contexts, or longitudinal studies, computational approaches could augment traditional analysis by efficiently identifying thematic patterns while preserving opportunities for human interpretation.

This paper presents a systematic, reproducible framework for applying BERTopic~\cite{grootendorst2022}, a neural topic model, to focus group discussion transcripts. Using data from ten focus groups exploring HPV vaccine perceptions in Tunisia, we address four methodological objectives: (1) systematic exploration of hyperparameter configurations to understand their influence on model outcomes, (2) assessment of topic stability through bootstrap resampling to quantify robustness, (3) comparison with Latent Dirichlet Allocation to contextualize BERTopic's performance against traditional approaches, and (4) validation of topic interpretability through independent human evaluation by domain experts. While our application focuses on health discourse, the methodological framework and findings have broader implications for topic modeling of small, heterogeneous conversational corpora common across qualitative research domains.

Our work makes three primary contributions to computational text analysis methodology. First, we develop and validate the first systematic framework specifically designed for topic modeling of focus group transcripts, demonstrating that transformer-based approaches can extract interpretable themes from conversational qualitative data when properly configured and validated. Second, we show that BERTopic substantially outperforms LDA for this data type, achieving 18\% higher coherence while discovering more interpretable topics, with human validation confirming that computational coherence metrics successfully predict expert judgments. Third, through transparent documentation of all modeling decisions, systematic evaluation across multiple quality dimensions, and rigorous comparison with traditional approaches, we contribute to the growing literature on reproducible practices in  qualitative research.

To ensure full reproducibility, we provide complete code, evaluation scripts, hyperparameter configurations, and detailed documentation of all modeling decisions. Code and anonymized data are available at ....

\section{Related Work}
\label{sec:relatedwork}

\subsection{Topic Modeling Approaches}

Topic modeling encompasses a family of statistical methods for discovering latent thematic structures in document collections. Latent Dirichlet Allocation (LDA)~\cite{blei2003} established the foundational probabilistic framework for topic modeling, modeling documents as mixtures of topics and topics as distributions over words. While LDA remains widely used~\cite{jelodar2018}, its bag-of-words assumption discards word order and contextual information.

BERTopic~\cite{grootendorst2022} represents a neural alternative, combining transformer-based sentence embeddings with UMAP dimensionality reduction~\cite{mcinnes2020}, HDBSCAN clustering~\cite{campello2013}, and class-based TF-IDF weighting. Comparative studies show BERTopic produces competitive coherence relative to LDA~\cite{egger2022}, though performance depends significantly on hyperparameter choices. Recent work~\cite{borcin_optimizing_2024} demonstrated that optimal configurations vary with corpus characteristics, underscoring the importance of systematic parameter exploration.

\subsection{Stability and Quality Assessment}

Assessing topic model quality requires evaluating multiple dimensions, as different metrics capture distinct aspects of model performance. Greene et al.~\cite{greene2014} advocated for bootstrap stability analysis as a criterion for topic model selection, proposing that models producing stable partitions across resampled data represent more reliable thematic structures. Stability is typically quantified using partition similarity measures such as Normalized Mutual Information (NMI) or Adjusted Rand Index (ARI)~\cite{hubert1985, strehl2002}. However, these metrics measure fundamentally different aspects of clustering agreement. ARI quantifies exact pairwise assignment agreement and is sensitive to cluster boundary variations~\cite{hubert1985}, while NMI measures information structure preservation based on entropy~\cite{strehl2002}. Recent systematic evaluations have demonstrated that NMI and ARI can produce different algorithm rankings depending on corpus characteristics~\cite{rudiger_topic_2022}, indicating that neither metric is universally superior and selection should align with analytical goals.

Semantic coherence, typically measured using the $C_v$ metric~\cite{roder_exploring_2015}, assesses whether topic words form interpretable semantic groups. However, coherence and stability represent distinct and sometimes conflicting quality dimensions~\cite{hosseiny2023, koltcov2024}. Koltcov et al.~\cite{koltcov2024} demonstrated empirically that models can exhibit high coherence while maintaining poor stability, underscoring the importance of evaluating both dimensions. Furthermore, algorithmic metrics frequently conflict with human judgments of topic quality~\cite{hoyle2021, chang2009}, and different coherence measures can disagree on model rankings~\cite{aletras2013}. Meaney et al.~\cite{meaney2023} demonstrate that different evaluation metrics often favor models of different complexity, indicating that no single metric captures all aspects of topic quality. These findings underscore the importance of multi-faceted evaluation that includes expert validation alongside computational metrics.

\subsection{Applications to Qualitative Research and Reproducibility Challenges}

Topic modeling has found increasing application in qualitative research across multiple domains. Substantial work has applied these methods to social media health discourse, including COVID-19 discussions on Twitter~\cite{qorib_covid-19_2023} and HPV vaccination debates~\cite{du_use_2020}. More recently, researchers have analyzed structured qualitative data including patient experience surveys~\cite{steele2025}, psychotherapy transcripts~\cite{lalk2024}, interview data~\cite{he2024}, and open-ended survey responses~\cite{guetterman_augmenting_2018, chang_accelerating_2021}. These applications demonstrate topic modeling's potential for efficiently identifying thematic patterns in qualitative data while highlighting methodological challenges including domain-informed preprocessing, handling conversational dynamics, and hierarchical topic organization~\cite{dou_hierarchicaltopics_2013}.

However, reproducibility has emerged as a critical concern. Hagg et al.~\cite{hagg2022} conducted a scoping review of 68 LDA studies in psychological science and found that most provided insufficient methodological detail to reproduce analyses, and that results are highly sensitive to analytic decisions. This challenge is particularly acute for small, heterogeneous qualitative corpora where configuration choices can profoundly affect outcomes. The lack of standardized evaluation protocols and transparent reporting practices undermines confidence in topic modeling findings, especially when results inform practical decisions. These issues highlight the need for systematic hyperparameter exploration, multi-dimensional quality assessment, rigorous validation protocols, and transparent documentation of all modeling choices.

Despite growing applications, we could find no prior work explicitly applying topic modeling to focus group discussion transcripts. This gap is notable: focus groups generate multi-party conversations where meaning emerges through interactive dialogue, differing substantively from both social media (spontaneous, brief posts) and individual interviews (single-voice narratives). The present study addresses this gap by developing and validating a systematic BERTopic framework for focus group analysis, with transparent reporting of all methodological decisions to support reproducibility and adaptation by other researchers.

\section{Methodology}
\label{sec:methodology}

\begin{figure}
    \centering
    \includegraphics[width=0.9\linewidth]{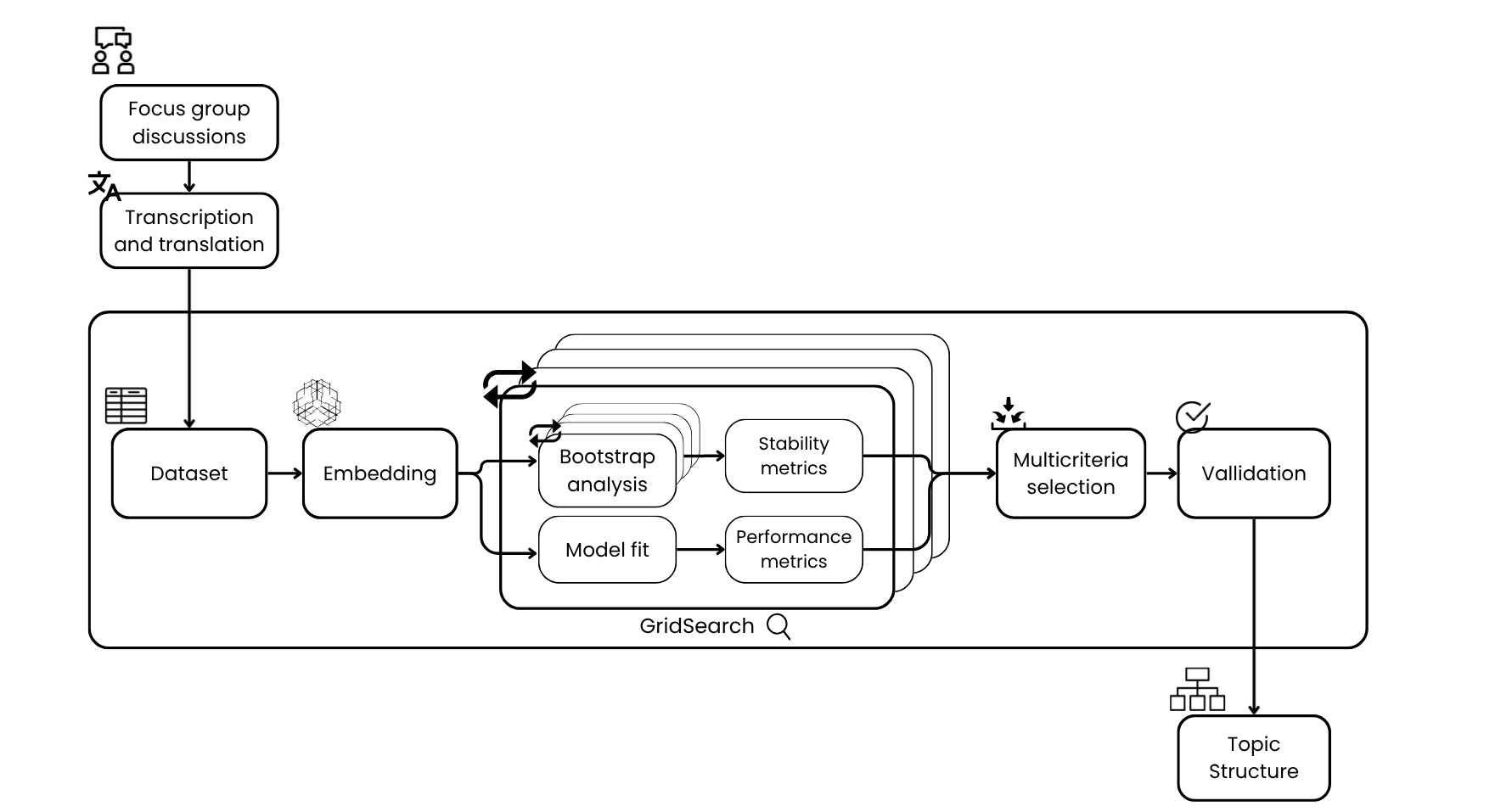}
    \caption{Systematic pipeline for topic modeling of focus group transcripts. The workflow begins with transcription and translation, followed by embedding generation. A grid search explores the chosen hyperparameter configurations, evaluating each through bootstrap stability analysis (NMI, ARI) and performance metrics (coherence, topic count, outlier fraction). Multicriteria selection balances performance metrics, with final validation by domain experts giving the interpretable topic structure.}
    \label{fig:pipeline}
\end{figure}

We apply BERTopic to HPV vaccine perception focus groups conducted in Tunisia. Our pipeline addresses three objectives: (1) systematic hyperparameter exploration, (2) stability assessment through bootstrap resampling, and (3) validation of topic interpretability. To contextualize BERTopic's performance, we also implement a tuned LDA baseline for comparison (detailed in Section~\ref{sec:lda_baseline}). Figure \ref{fig:pipeline} illustrates our methodology pipeline.

\subsection{Data collection and description}
Between September 2023 and February 2024, we conducted ten focus group discussions exploring attitudes and perceptions toward HPV vaccination in Tunisia. The study received approval from the Institutional Review Board of Pasteur Institute of Tunis. Participants provided written informed consent, and all discussions were audio recorded and transcribed from Tunisian Dialect to French. Participants were community representatives: individuals in daily contact with children and involved in health-related decision-making, such as parents, teachers, and childcare professionals. To account for socioeconomic and cultural disparities, participants were recruited across five geographically diverse regions, with two gender-stratified sessions per region. Discussions averaged 120 minutes, were conducted in Tunisian dialect, and were moderated by a socio-anthropologist and a medical doctor using a semi-structured guide. 

All discussions were audio recorded, transcribed, translated to French, and parsed into discrete rows representing individual speaker turns. Each turn was annotated with participant identifiers and enriched with metadata including gender, region, and age. Moderator questions and introductory statements were excluded to focus the analysis on substantive participant perspectives. Text preprocessing was kept minimal to preserve semantic content: we applied basic cleaning, standardized capitalization, and normalized equivalent terms (e.g., papillomavirus/HPV, coronavirus/COVID), but retained stop words to maintain contextual information for transformer-based embeddings. We removed utterances whose length was below the 5th or above the 99th percentile to remove entries that either contained too little information to be meaningfully embedded or were excessively long due to digressions or transcription artifacts. 

The resulting corpus contained 1,075 speaker turns or utterances, with a mean length of 32.7 words (SD = 25.0).  Table~\ref{tab:sample_data} presents a sample of annotated and translated speaker turns.

\begin{table}
\centering
\scalebox{0.83}{
\begin{tabular}{p{7cm} p{7cm} l l r}
\toprule
\textbf{Original (French)} & \textbf{Translation (English)} & \textbf{ParticipantID} & \textbf{Region} & \textbf{Age} \\
\midrule
En vérité, les centres de santé de base sont toujours surpeuplés et il y a des dizaines de patients quotidiennement qui ... & In fact, basic health centers are always overcrowded, with dozens of patients coming in every day... & par97339 & Region1 & 45 \\
Pardon, mais d'après mon expérience personnelle et ce que j'entends, la majorité des gens ont eu des troubles de mémoire... & Excuse me, but from my personal experience and what I hear, most people have had memory problems... & par39944 & Region2 & 47 \\
Lorsqu'une femme souffre d'un cancer du col de l'utérus, elle préfère ne le dire à personne. Elle reste silencieuse, car... & When a woman suffers from cervical cancer, she prefers not to tell anyone. She remains silent because... & par73108 & Region3 & 43 \\
Pourquoi ne vaccinons-nous que les filles et pas les garçons .... & Why do we vaccinate only girls and not boys?... & par83355 & Region4 & 38 \\
Nous sommes pour l'obligation de vaccination.... & We are in favor of mandatory vaccination... & par03182 & Region3 & 22 \\
Je me suis disputée avec mon mari à cause du vaccin. Il travaille à l'hôpital et a essayé de me convaincre quotidienneme... & I argued with my husband about the vaccine. He works at the hospital and tried to convince me every day... & par49154 & Region1 & -- \\
\bottomrule
\end{tabular}
}
\vspace{8pt}
\caption{Sample of annotated and translated speaker turns (utterances). Participants names and regions have been anonymized and the dataset doesn't contain any Personally Identifiable Information}
\label{tab:sample_data}
\end{table}

\subsection{Embedding Generation and BERTopic Implementation Pipeline}
We used the \texttt{sentence-transformers/paraphrase-multilingual-mpnet-base-v2} model\cite{reimers2019} to generate 768-dimensional sentence embeddings for all preprocessed sentences. This model is multilingual and optimized for sentence semantic similarity tasks, making it relevant for topic modeling in French.

Importantly, the embeddings were computed once and stored for use across all subsequent model configurations, ensuring that any observed differences in topic quality or stability result from hyperparameter variations rather than inconsistencies in semantic representation. This also provides computational efficiency by eliminating redundant embedding calculations during hyperparameter exploration.

We implemented BERTopic with its standard three-component architecture: UMAP for dimensionality reduction, HDBSCAN for density-based clustering, and class-based TF-IDF for topic representation. The pipeline was configured for French language processing to ensure appropriate stop-word filtering and linguistic handling.

\subsection{Hyperparameter Exploration and Stability Analysis}
We conducted systematic hyperparameter exploration across three key parameters: UMAP \texttt{n\_neighbors}, UMAP \texttt{n\_components}, and HDBSCAN \texttt{min\_cluster\_size}. The parameter ranges were selected to span from fine-grained to coarser topic discovery: \texttt{n\_neighbors} $\in \{10, 15, 30\}$, \texttt{n\_components} $\in \{5, 10, 50\}$, and \texttt{min\_cluster\_size} $\in \{5, 10, 15\}$. To control stochasticity and ensure reproducibility, we set a global random seed that propagates through all random components of the pipeline.

For each of the 27 resulting parameter combinations, we performed the following evaluation protocol:

\textbf{Bootstrap Stability Analysis:} We generated 30 bootstrap replicates by sampling 85\% of the original sentences with replacement. This sampling fraction provides sufficient data for reliable topic discovery while introducing meaningful variation to assess model robustness. Each bootstrap replicate used a deterministic random seed (base seed + replicate number) to ensure reproducible sampling. For each bootstrap replicate, we instantiated fresh UMAP and HDBSCAN models. This is essential because fitted dimensionality reduction and clustering models retain internal state that could bias results if reused across different data samples. Each bootstrap model was fitted independently on its respective data sample. 

We assessed model stability using Normalized Mutual Information (NMI) and the Adjusted Rand Index (ARI), computed between all pairs of bootstrap replicates. For each metric, pairwise scores were computed only for documents appearing in both replicates, and the median value across all pairwise comparisons was retained as the stability estimate for that parameter configuration.

\textbf{Full Dataset Evaluation:} We fitted a BERTopic model on the complete dataset using each parameter configuration. For each fitted model, we computed topic coherence using $C_v$ coherence scores, recorded the number of discovered topics, and calculated the proportion of documents assigned to the outlier category.

\subsection{Multi-Criteria Model Selection}

Rather than applying rigid optimization criteria or automated selection rules, we adopted a multi-criteria decision framework balancing four dimensions:

\begin{itemize}
\item \textbf{Coherence}: Semantic interpretability of topics, measured by $C_v$ scores
\item \textbf{Stability}: Bootstrap consistency, assessed through NMI and ARI
\item \textbf{Granularity}: Number of non-outlier topics, determining thematic resolution
\item \textbf{Coverage}: Proportion of documents assigned to topics versus outliers
\end{itemize}

This approach acknowledges that optimal configurations depend on research context: exploratory qualitative research may prioritize coherence, while studies requiring reproducible classifications emphasize stability. By transparently presenting performance across all 27 configurations, we enable informed selection rather than imposing a single optimization objective.


\subsection{Human Validation}

To assess the coherence and interpretability of our final topic model, we conducted a validation study with three independent annotators having expertise in public health and qualitative analysis. Each annotator rated the coherence of each topic on a 5-point Likert scale (1=Incoherent, 5=Highly Coherent) based on 10 representative documents randomly selected from each topic. Raters were blind to topic labels and model-generated keywords, and first evaluated coherence independently before collaboratively assigning interpretable labels to each topic.

Inter-annotator agreement was assessed using weighted Cohen's Kappa\cite{mchugh_interrater_2012}, which accounts for the ordinal nature of ratings, and the Intraclass Correlation Coefficient (ICC(2,1))\cite{shrout_intraclass_1979}, which quantifies overall reliability across all three raters.

\subsection{LDA Baseline Comparison}
\label{sec:lda_baseline}

To contextualize the performance of our BERTopic approach, we implemented Latent Dirichlet Allocation as a comparative baseline. Following current practice in applied topic modeling research~\cite{hagg2022}, we tuned only the number of topics while maintaining default settings for other hyperparameters. Hagg et al.'s scoping review found that most applications leave parameters at defaults, making this approach representative of typical usage.

We performed grid search across the number of topics $\in \{3,\cdots , 15\}$, selected to span coarse to fine-grained granularity appropriate for our corpus size. Each configuration was evaluated using the same $C_v$ coherence scoring framework applied to BERTopic models. The optimal LDA configuration was selected based on maximum coherence score.

\section{Results}
\label{sec:results}

\subsection{Stability Analysis}

We evaluated topic model stability across 27 hyperparameter combinations using bootstrap resampling with 30 replicates per configuration. This analysis showed NMI values ranging from 0.388 to 0.703 (median: 0.558), while ARI values ranged from 0.107 to 0.202 (median: 0.156). Figure~\ref{fig:gridsearch_metrics} (top panels) presents these stability metrics across the parameter space.

The two metrics reveal contrasting patterns. NMI values are highest for \texttt{min\_cluster\_size} = 5 configurations (0.63-0.70), decreasing substantially for larger cluster sizes: \texttt{min\_cluster\_size} = 10 achieves 0.48-0.56, while \texttt{min\_cluster\_size} = 15 shows 0.39-0.44. ARI values are systematically lower and exhibit the opposite trend: configurations with \texttt{min\_cluster\_size} = 15 achieve the highest ARI scores (0.15-0.20), while \texttt{min\_cluster\_size} = 5 shows the lowest values (0.11-0.16).

This divergence reflects fundamental differences in what these metrics quantify. NMI measures information structure preservation and remains stable when fine-grained clusters consistently capture semantic distinctions, even when exact boundaries shift across bootstrap samples. ARI measures exact pairwise assignment agreement and is highly sensitive to boundary variations, particularly in small clusters where minor document shifts represent large proportional changes. The negative correlation between NMI and ARI (r = -0.691) further illustrates their divergent behavior.

Critically, we observed a strong negative correlation between NMI and model coherence (r = -0.911, p < 0.001). Configurations with highest NMI (0.63-0.70) produced the least coherent topics (coherence = 0.39-0.43), while configurations with lower NMI (0.39-0.44) achieved substantially higher coherence (0.52-0.59). In contrast, ARI demonstrated a moderate positive correlation with coherence (r = 0.639, p < 0.001), though the universally low ARI range limits its practical utility for model selection. These patterns indicate that stability as measured by NMI and semantic quality as measured by coherence capture fundamentally different constructs, with implications for model selection strategies.

\subsection{Performance Assessment}

\begin{figure}[htbp]
\centering
\includegraphics[width=0.85\textwidth]{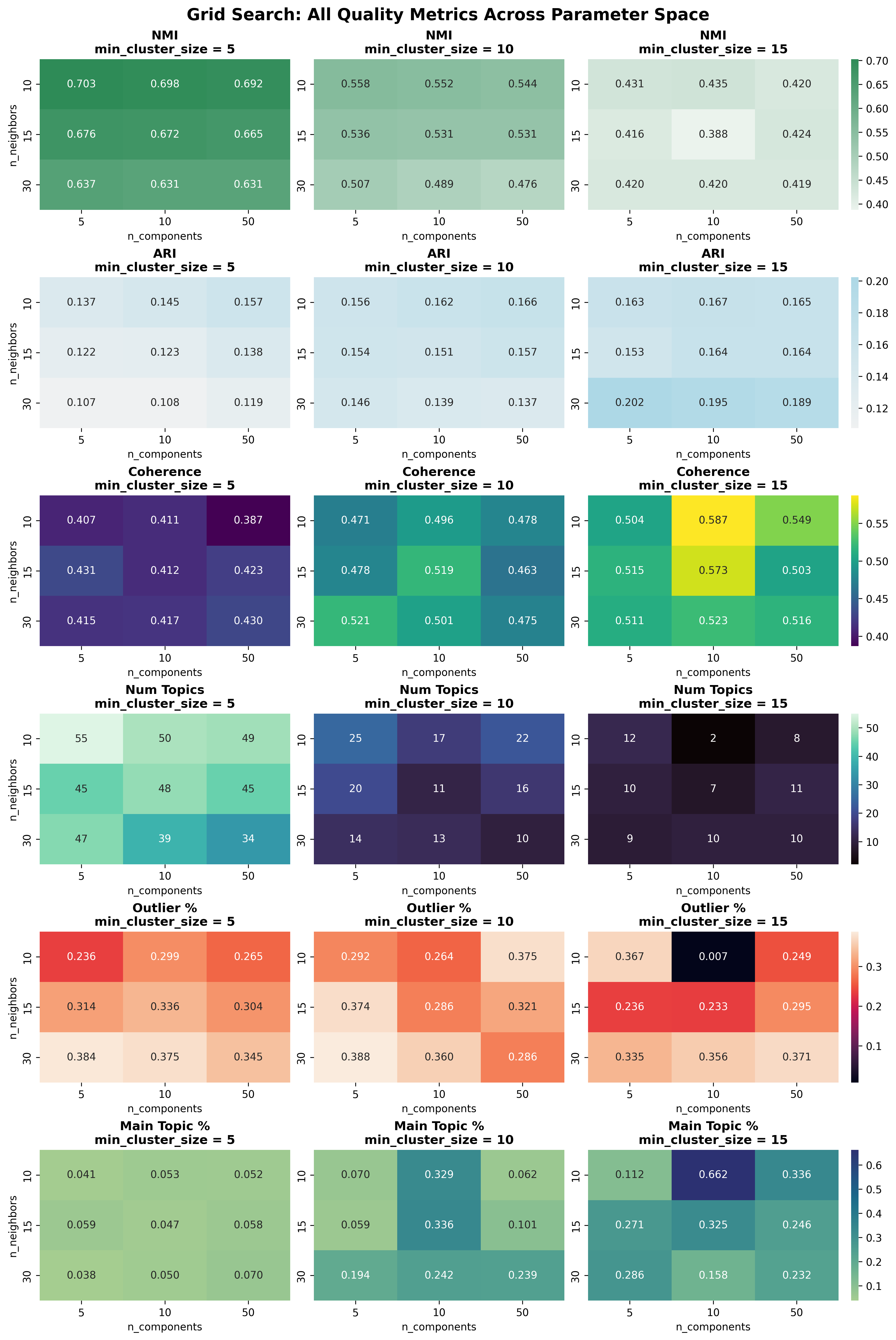}
\caption{Quality metrics across the complete hyperparameter space. Six metrics are displayed as heatmaps across three minimum cluster sizes (columns). Each heatmap shows UMAP \texttt{n\_neighbors} (rows: 10, 15, 30) by UMAP \texttt{n\_components} (columns: 5, 10, 50). Top panels show bootstrap stability (NMI and ARI), middle panels show topic quality (coherence and number of topics), and bottom panels show coverage metrics (outlier fraction and main topic proportion). }
\label{fig:gridsearch_metrics}
\end{figure}

Figure~\ref{fig:gridsearch_metrics} (middle and bottom panels) displays performance metrics across the entire parameter space. Coherence scores vary substantially, ranging from 0.387 to 0.587 (mean = 0.478, SD = 0.060). The highest coherence is achieved by coarser configurations with \texttt{min\_cluster\_size} = 15 (0.50-0.59), while \texttt{min\_cluster\_size} = 5 configurations show lower coherence (0.39-0.43) despite higher stability by NMI.

The number of discovered topics varies dramatically with \texttt{min\_cluster\_size}: configurations with \texttt{min\_cluster\_size} = 5 produce 34-55 topics, \texttt{min\_cluster\_size = 10} yields 10-25 topics, and \texttt{min\_cluster\_size = 15} generates 2-12 topics. Outlier fractions range from 0.7\% to 38.8\%, with generally higher outlier rates for finer-grained configurations. Main topic proportions range from 4\% to 66\%, indicating substantial variation in topic size distributions.


\subsection{Model Selection}

Table~\ref{tab:top_configs} presents the top 10 configurations ranked by coherence score. 

\begin{table}[htbp]
\centering
\small
\begin{tabular}{ccccccccc}
\toprule
\textbf{Rank} & \textbf{nn} & \textbf{nc} & \textbf{mcs} & \textbf{Topics} & \textbf{Coherence} & \textbf{NMI} & \textbf{ARI} & \textbf{Outliers} \\
\midrule
1 & 10 & 10 & 15 & 2 & 0.587 & 0.435 & 0.167 & 0.007 \\
2* & 15 & 10 & 15 & 7 & 0.573 & 0.388 & 0.164 & 0.233 \\
3 & 10 & 50 & 15 & 8 & 0.549 & 0.420 & 0.165 & 0.249 \\
4 & 30 & 10 & 15 & 10 & 0.523 & 0.420 & 0.195 & 0.356 \\
5 & 30 & 5 & 10 & 14 & 0.521 & 0.507 & 0.146 & 0.388 \\
6 & 15 & 10 & 10 & 11 & 0.519 & 0.531 & 0.151 & 0.286 \\
7 & 30 & 50 & 15 & 10 & 0.516 & 0.419 & 0.189 & 0.371 \\
8 & 15 & 5 & 15 & 10 & 0.515 & 0.416 & 0.153 & 0.236 \\
9 & 30 & 5 & 15 & 9 & 0.511 & 0.420 & 0.202 & 0.335 \\
10 & 10 & 5 & 15 & 12 & 0.504 & 0.431 & 0.163 & 0.367 \\
\bottomrule
\end{tabular}
\vspace{8pt}
\caption{Top 10 configurations ranked by coherence score. Configurations are identified by base hyperparameters (\texttt{n\_neighbors} (nn), \texttt{n\_components}(nc), and \texttt{min\_cluster\_size} (mcs)). The selected configuration (rank 2, marked with *) balances high coherence with reasonable granularity.}
\label{tab:top_configs}
\end{table}

Based on our multicriteria selection approach,
we selected the configuration ranking second by coherence: \texttt{n\_neighbors} = 15, \texttt{n\_components} = 10, \texttt{min\_cluster\_size = 15}, which discovered 7 non-outlier topics. This configuration achieved coherence of 0.573 with bootstrap stability of NMI = 0.388 and ARI = 0.164, assigning 23.3\% of documents to outliers with the largest topic containing 32.5\% of non-outlier documents.

The selection rationale balanced multiple criteria: while rank 1 achieved marginally higher coherence (0.587), its collapse to only 2 topics was deemed too coarse for meaningful exploration of vaccine perceptions. Our selected model provides sufficient granularity (7 topics) to capture distinct thematic dimensions while maintaining strong coherence. The moderate stability metrics reflect the broader pattern that high-coherence models show lower NMI due to coarser clustering, consistent with the strong negative correlation documented above.

\subsection{LDA Baseline Comparison}

We compared our BERTopic model against Latent Dirichlet Allocation using the same coherence-based evaluation framework with a grid search across 3 to 15 topics, and evaluated each configuration using C$_v$ coherence.

LDA coherence scores ranged from 0.366 (15 topics) to 0.486 (3 topics), with performance generally declining as granularity increased. The optimal configuration identified 3 topics with coherence of 0.486. Table~\ref{tab:lda_comparison} compares the optimized LDA baseline with our BERTopic model.

\begin{table}[htbp]
\centering
\begin{tabular}{lccc}
\toprule
\textbf{Model} & \textbf{Topics} & \textbf{Coherence} & \textbf{Outliers (\%)} \\
\midrule
LDA (optimized) & 3 & 0.486 & N/A \\
BERTopic (selected) & 7 & 0.573 & 23.3 \\
\midrule
\multicolumn{2}{l}{\textit{Improvement:}} & +18\% & -- \\
\bottomrule
\end{tabular}
\vspace{8pt}
\caption{Comparison of optimized LDA baseline and final BERTopic model. BERTopic achieves 18\% higher coherence while discovering more than twice as many interpretable topics.}
\label{tab:lda_comparison}
\end{table}

Our BERTopic model substantially outperformed the baseline, achieving 18\% higher coherence (0.573 vs. 0.486) while discovering more than twice as many interpretable topics (7 vs. 3). This performance gap likely reflects architectural differences: LDA's bag-of-words assumptions treat documents as unordered token collections, while BERTopic's transformer-based embeddings capture contextual semantics. For conversational data where meaning depends on discourse context (such as ``je n'ai pas confiance'' referencing different trust domains across utterances) contextualized representations appear better suited to capturing semantic nuance. The LDA baseline's optimal configuration at only 3 topics suggests that without contextualized embeddings, the model struggles to differentiate finer semantic distinctions in French-language focus group discourse.

\subsection{Final Topic Structure and Human Validation}

Human validation by three independent domain experts confirmed topic coherence and interpretability. The seven emerging themes as labeled by the annotators were: General health knowledge and taboos (Topic 0, 32\%), Vaccination importance and institutional processes (Topic 1, 23\%), Trusted information sources (Topic 2, 9\%), Personal vaccination experiences (Topic 3, 5\%), Public versus private healthcare debate (Topic 4, 3\%), COVID-19 vaccine adverse effects (Topic 5, 3\%), and Institutional trust erosion post-pandemic (Topic 6, 1\%). 

Inter-annotator agreement was robust: ICC(2,1) = 0.700 indicates very good reliability per ~\cite{koo_guideline_2016}, while mean weighted Cohen's kappa of 0.678 demonstrates substantial agreement~\cite{landis1977}. Annotators achieved perfect agreement within one-point tolerance for 85.7\% of topics (6/7).

Table~\ref{tab:human_validation} presents coherence ratings for each topic. Five of seven topics received mean ratings $\geq$ 4.0, indicating good-to-excellent coherence. Ratings ranged from 1.67 to 4.67 (overall mean = 3.81). Topic 0 received the lowest rating (mean = 1.67), reflecting its broad, heterogeneous content spanning multiple health domains. Topics 1, 2, 3, 5, and 6 achieved mean ratings between 4.0 and 4.67, confirming strong semantic coherence. Topic 4 received moderate ratings (mean = 3.33). Topic size showed no correlation with coherence ratings ($\rho$ = -0.213, p = 0.583), confirming evaluations reflected semantic quality rather than prevalence.

\begin{table}[htbp]
\centering
\begin{tabular}{cccccl}
\toprule
\textbf{Topic ID} & \textbf{Ann. 1} & \textbf{Ann. 2} & \textbf{Ann. 3} & \textbf{Mean} & \textbf{Topic Label}  \\
\midrule
0 & 2 & 2 & 1 & 1.67 & General health knowledge and taboos \\
1 & 5 & 5 & 3 & 4.33 & Vaccination importance and processes \\
2 & 5 & 5 & 4 & 4.67 & Trusted information sources \\
3 & 4 & 5 & 4 & 4.33 & Personal vaccination experiences \\
4 & 3 & 4 & 3 & 3.33 & Public vs. private healthcare debate \\
5 & 4 & 5 & 4 & 4.33 & COVID-19 vaccine adverse effects \\
6 & 4 & 4 & 4 & 4.00 & Erosion of institutional trust post-COVID \\
\midrule
\multicolumn{6}{l}{\textit{Inter-rater reliability: Weighted Cohen's $\kappa$ = 0.678; ICC(2,1) = 0.700}} \\
\bottomrule
\end{tabular}
\vspace{8pt}
\caption{Human validation results. Coherence scores range from 1 (incoherent) to 5 (highly coherent).}
\label{tab:human_validation}
\end{table}

\begin{figure}[ht]
\centering
\includegraphics[width=1\textwidth]{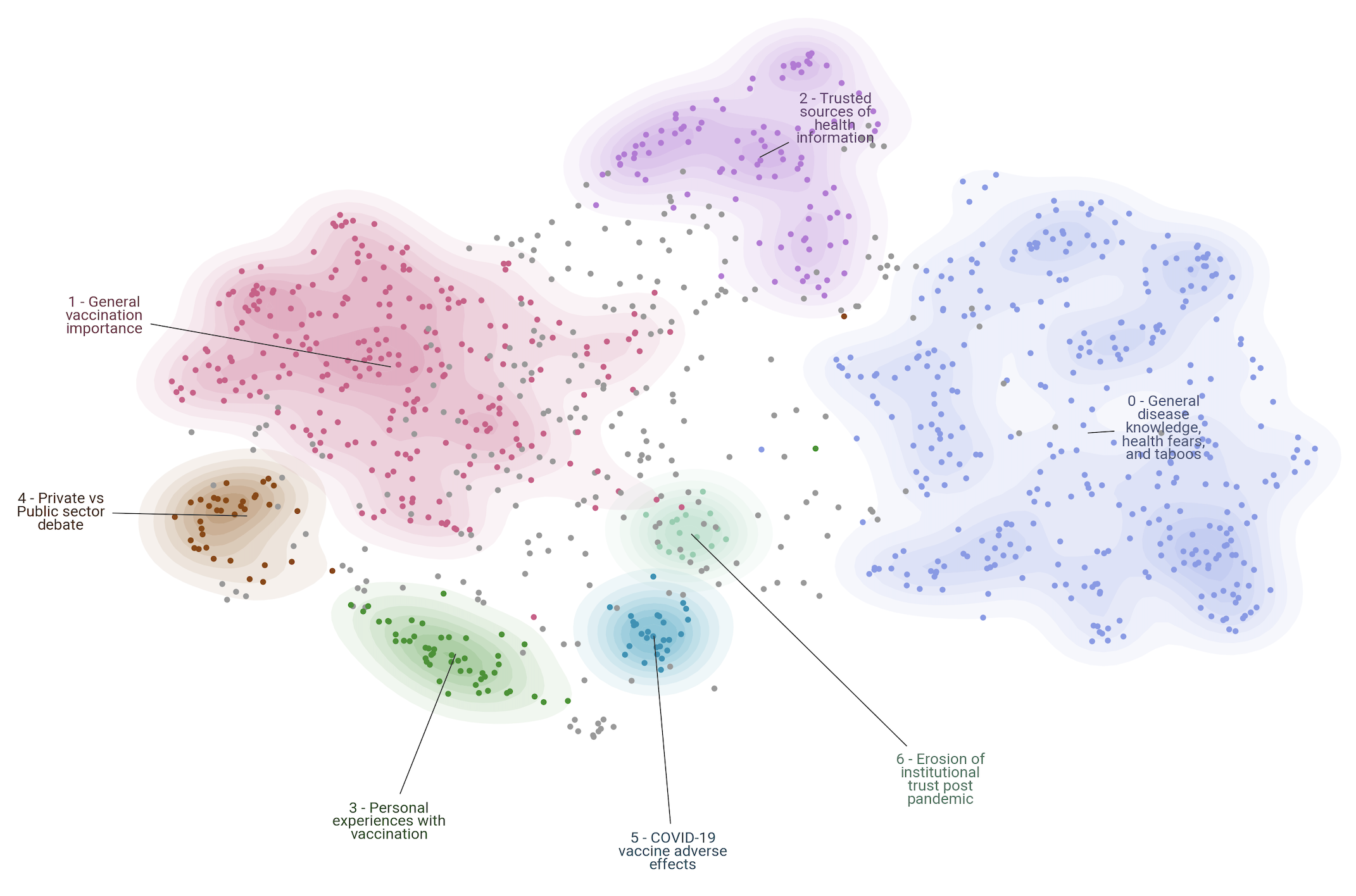}
\caption{Two-dimensional visualization of document embeddings colored by final merged topic assignments. Each point represents a document (utterance) from the focus group transcripts. Colored regions show the 7 merged topics, with spatial proximity indicating semantic similarity.}
\label{fig:document_datamap}
\end{figure}

Figure~\ref{fig:document_datamap} visualizes the semantic organization of the topics in two-dimensional embedding space, revealing distinct thematic clusters and their spatial relationships. While this paper focuses on methodological validation, the substantive themes identified merit deeper analysis to understand their public health implications. A forthcoming companion paper will provide detailed thematic analysis and discuss policy recommendations based on these findings.

\section{Discussion}
\label{sec:discussion}

\subsection{Bootstrap Stability: A Multifaceted Construct}

Our bootstrap stability analysis revealed that partition similarity metrics capture fundamentally different aspects of clustering agreement. NMI and ARI exhibited a strong negative correlation (r = -0.691), systematically disagreeing on which configurations represent stable models. NMI favored fine-grained configurations (min\_cluster\_size = 5) producing 34-55 topics with high information structure preservation (NMI = 0.63-0.70), while ARI favored coarser configurations (min\_cluster\_size = 15) with fewer topics where exact pairwise assignments remained more consistent (ARI = 0.15-0.20).

This divergence reflects fundamental differences in what these metrics quantify. NMI measures whether clustering solutions preserve the same information structure based on entropy, remaining stable when fine-grained semantic distinctions are consistently recovered across bootstrap samples even if exact cluster boundaries shift~\cite{strehl2002}. ARI measures exact pairwise assignment agreement, highly sensitive to boundary variations particularly in small clusters where minor document reassignments represent large proportional changes~\cite{hubert1985}. Recent systematic evaluations have confirmed that NMI and ARI can produce different algorithm rankings depending on corpus characteristics~\cite{rudiger_topic_2022}, indicating that "stability" is not a monolithic construct but depends on which aspect of partition agreement is prioritized.

Moreover, these stability metrics showed different relationships with topic interpretability. NMI exhibited a strong negative correlation with coherence (r = -0.911), while ARI showed moderate positive correlation (r = 0.639) though with limited discriminative power due to universally low values. This metric-dependent relationship demonstrates that bootstrap stability and semantic quality represent distinct dimensions that may conflict rather than align. These findings suggest that stability assessment should employ multiple metrics to capture different facets of partition robustness, and that stability measures alone provide insufficient guidance for model selection in exploratory research contexts where interpretability is paramount.

\subsection{Multi-Criteria Model Selection and Validation}

The competing objectives revealed by our systematic evaluation (possible stability-coherence tradeoff, varying topic granularity, and different coverage patterns) demonstrate why rigid optimization strategies prove insufficient in practice. While the highest-coherence configuration collapsed to only 2 topics, our selection of 7 topics reflects a judgment that thematic resolution matters alongside semantic quality. This decision acknowledges that optimal configurations depend on research goals: classification tasks might prioritize stability for reproducibility, while exploratory analysis requires balancing interpretability with analytical utility.

Human validation by three domain experts confirmed that coherence-prioritized selection produced interpretable topics, with robust inter-annotator agreement (ICC = 0.700, $\kappa$ = 0.678) and the majority of topics receiving high ratings. Importantly, ratings reflected semantic quality rather than topic prevalence, confirming that computational coherence metrics successfully predict expert judgments of interpretability in our application domain.

The alignment between automated C$_v$ scores and human assessments addresses longstanding concerns that algorithmic optimization may produce statistically optimal but substantively incoherent topics~\cite{chang2009}. Our multi-criteria framework provides a transparent, defensible alternative to single-metric optimization or default configurations.

\subsection{Comparison with LDA Baseline}

The substantial performance gap between BERTopic and LDA (18\% higher coherence, 7 vs. 3 topics) likely reflects fundamental architectural differences. LDA's bag-of-words assumptions treat documents as unordered token collections, relying on statistical co-occurrence patterns to infer latent topics. In contrast, BERTopic's transformer-based sentence embeddings capture contextual semantics, preserving meaning that depends on word order and discourse context.

For conversational focus group data, this distinction proves particularly important. The same phrase may reference different concepts depending on surrounding discourse. Contextual embeddings can distinguish these semantic nuances, while bag-of-words representations collapse them into undifferentiated co-occurrence statistics. The LDA baseline's optimal configuration at only 3 topics suggests that without contextualized representations, the model struggles to differentiate finer semantic distinctions, producing overly broad topics that sacrifice thematic resolution for coherence.

However, our comparison has limitations. Following standard practice~\cite{hagg2022}, we tuned only the number of topics for LDA while maintaining default hyperparameters for Dirichlet priors and learning parameters. More extensive hyperparameter exploration could potentially improve LDA performance, though we consider our approach representative of typical applied usage. Additionally, we did not compare against other neural topic modeling approaches such as Top2Vec or CTM, limiting our ability to assess BERTopic's performance relative to alternative contextualized methods.

\subsection{Limitations and Future Directions}

Several limitations merit consideration. First, our study is restricted to French-language focus group transcripts from Tunisia. The multilingual embedding model we employed may perform differently across languages, and the identified topic structures likely reflect culturally specific patterns of health discourse. Validation in other linguistic and cultural contexts is needed to assess the broader applicability of our methodological framework.

Second, the substantial sensitivity of results to hyperparameter choices raises questions about generalizability. While our systematic exploration represents a more rigorous approach than accepting default settings, specific optimal configurations may not transfer to focus groups with different conversational dynamics or substantive domains.

Third, one large topic received poor coherence ratings from evaluators, indicating the model produced a heterogeneous catch-all category rather than a unified theme. This suggests limitations in BERTopic's density-based clustering for highly diverse discourse, where some utterances may not fit coherent thematic patterns. Finally, we compared only against LDA rather than other neural methods, and our LDA baseline was tuned only on number of topics following standard practice. More comprehensive benchmarking across methods and fuller hyperparameter exploration would strengthen comparative conclusions.

Future work should examine whether the stability-coherence tradeoff generalizes across languages and data types, explore alternative stability metrics that better align with interpretability, and conduct systematic comparisons with other neural topic modeling approaches. Integration with manual qualitative coding frameworks could also clarify how computational topics relate to human-identified themes.

\subsection{Implications for Focus Group Research}

This work demonstrates the feasibility and value of applying systematic topic modeling to focus group transcripts. Focus groups remain a cornerstone methodology in qualitative health research, yet the labor-intensive nature of traditional manual coding limits the scale of synthesis possible across multiple sessions. Computational approaches do not replace human interpretation but can augment it by efficiently identifying thematic patterns, particularly valuable for large-scale focus group studies where manual analysis of hundreds of transcripts becomes prohibitive.

Our findings suggest several practical recommendations for researchers considering topic modeling for focus group analysis. First, invest in systematic hyperparameter exploration rather than accepting default configurations, as performance varies dramatically depending on corpus characteristics. Second, evaluate models across multiple complementary metrics (stability, coherence, granularity, and coverage) and transparently document the tradeoffs inherent in final selection decisions. Third, validate computational outputs through independent human review by domain experts using structured protocols; we recommend at least three raters evaluating representative documents per topic on standardized coherence scales. Fourth, apply minimal preprocessing for transformer-based models, preserving stop words and contextual elements that bag-of-words approaches would discard. These practices balance computational efficiency with the interpretive rigor qualitative research demands, providing a principled framework for augmenting traditional focus group analysis with algorithmic support.

\section{Conclusion}
\label{sec:conclusion}

This paper presents a systematic, reproducible computational framework for applying BERTopic to focus group transcripts, contributing validated methodological practices to the qualitative research toolkit. Through rigorous exploration of 27 hyperparameter configurations with bootstrap stability assessment and formal human validation, we establish that transformer-based topic modeling can reliably extract interpretable themes from structured conversational data. The strong inter-rater agreement (ICC = 0.700) confirms that computational approaches, when properly validated, produce outputs aligned with expert domain understanding.

Our work addresses a critical gap: while topic modeling has been extensively applied to social media and online text, focus group discussions, a cornerstone of qualitative research, have received limited computational attention. By demonstrating feasibility on transcripts with naturally imbalanced topic distributions and relatively small corpus size (1,075 utterances), we extend neural topic modeling to structured research settings where traditional coding remains dominant. The transparent documentation of all modeling choices, evaluation procedures, and quality tradeoffs provides a reproducible template that researchers can adapt to their own focus group contexts.

This methodological foundation enables substantive analysis of vaccine perceptions in Tunisia, where our identified themes reveal specific patterns in health discourse, including vaccine hesitancy drivers, information source credibility, and trust in health institutions. These findings will inform targeted communication strategies and policy interventions in the forthcoming companion paper. More broadly, our framework supports the integration of computational efficiency with interpretive depth, allowing researchers to scale qualitative synthesis across multiple sessions while maintaining the contextual understanding that characterizes rigorous social research. As focus groups remain central to health research, market research, and social science inquiry, validated computational methods that respect conversational data characteristics hold significant potential for advancing both methodological practice and substantive understanding.

\appendix

\section{Appendix: Representative Topic Examples}
\label{app:topic_examples}

Table~\ref{tab:topics_appendix} provides representative utterances for each of the 7 topics identified by our BERTopic model, presented in both original French and English translation. These examples illustrate the thematic content and semantic coherence of each topic cluster.

\begin{landscape}
\begin{table}[p]
\centering
\begin{scriptsize}
\begin{tabular}{p{3.2cm} p{9.2cm} p{9.2cm}}

\toprule
\textbf{Topic Label} & \textbf{Original (French)} & \textbf{Translation (English)} \\
\midrule

\textbf{0 - General disease knowledge, health fears, and taboos} &
\begin{compactitem}
\item Je pense que c'est un sujet délicat et on ne peut pas en parler avec tout le monde.
\item Oui, exactement. Ma collègue m'a confié que ce qu'elle vivait dépassait ma compréhension. Par exemple, les douleurs qu'elle ressentait aux toilettes, les symptômes étranges...
\item C'est une maladie existante mais rare.
\end{compactitem} &
\begin{compactitem}
\item I think this is a sensitive topic, and you cannot talk about it with everyone.
\item Yes, exactly. My colleague confided that what she was going through was beyond my understanding. For example, the pain she felt when using the bathroom, the strange symptoms...
\item It is a real but rare disease.
\end{compactitem} \\

\midrule
\textbf{1 - General vaccination importance} &
\begin{compactitem}
\item Moi, je suis convaincue qu'il faut faire tous les vaccins, bien que je n'aie pas encore d'enfants.
\item La question de la vaccination obligatoire est importante depuis toujours, car les établissements scolaires et d'autres institutions exigent des élèves, étudiants et employés ce document, le certificat de vaccination.
\item La majorité d'entre nous suivent le carnet de vaccination.
\end{compactitem} &
\begin{compactitem}
\item I am convinced that one should take all vaccines, even though I do not yet have children.
\item The issue of mandatory vaccination has always been important because schools and other institutions require students and employees to provide this document: the vaccination certificate.
\item Most of us follow the vaccination schedule.
\end{compactitem} \\

\midrule
\textbf{2 - Trusted sources of health information} &
\begin{compactitem}
\item Je prends l'information de tout médecin ou spécialiste, qu'il travaille dans le secteur public ou privé.
\item Il ne faut pas dépendre des médecins. Chacun peut s'informer par ses propres moyens. Facebook est une source importante pour moi et j'ai confiance dans les informations publiées sur Facebook.
\item Je ne suis pas d'accord avec elle. Le ministère de la Santé n'est pas n'importe qui. Il faut suivre les instructions du ministère.
\end{compactitem} &
\begin{compactitem}
\item I take information from any doctor or specialist, whether they work in the public or the private sector.
\item You should not depend on doctors. Everyone can inform themselves on their own. Facebook is an important source for me, and I trust the information published there.
\item I do not agree with her. The Ministry of Health is not just anyone. You must follow the ministry’s instructions.
\end{compactitem} \\

\midrule
\textbf{3 - Personal experiences with vaccination} &
\begin{compactitem}
\item Moi, je ne l'ai pas reçu car j'étais enceinte à l'époque de la vaccination et j'ai attrapé le virus. J'ai beaucoup souffert et j'ai eu de gros problèmes respiratoires.
\item Je n'étais pas convaincue, mais j'ai dû me faire vacciner.
\item Le vaccin contre la grippe, je l'ai fait, c'est un peu cher, cette année je ne l'ai pas fait.
\end{compactitem} &
\begin{compactitem}
\item I did not receive it because I was pregnant at the time of vaccination, and I caught the virus. I suffered a lot and had severe respiratory problems.
\item I was not convinced, but I had to get vaccinated.
\item I got the flu vaccine; it is a bit expensive. This year, I did not get it.
\end{compactitem} \\

\midrule
\textbf{4 - Private vs Public sector debate} &
\begin{compactitem}
\item C'est injuste. Vous pensez que c'est normal que je paie cette somme pour un vaccin que j'ai normalement le droit d'obtenir gratuitement ! C'est mon droit !
\item Mais il y a plusieurs lacunes dans le secteur public qu'il faut reconnaître. Le secteur privé dans le domaine de la santé est devenu supérieur au secteur public...
\item Parce que ma fille a suivi sa grossesse chez un médecin privé et a accouché dans une clinique privée. Elle n'a donc pas eu recours au secteur public dès le début, choisissant de surveiller son enfant chez un pédiatre privé.
\end{compactitem} &
\begin{compactitem}
\item It is unfair. Do you think it is normal for me to pay this amount for a vaccine that I have the right to obtain for free? It is my right!
\item But several shortcomings in the public sector must be acknowledged. The private health sector has become superior to the public sector...
\item Because my daughter followed her pregnancy with a private doctor and gave birth in a private clinic. She therefore did not use the public sector from the beginning, choosing to monitor her child with a private pediatrician.
\end{compactitem} \\

\midrule
\textbf{5 - COVID-19 vaccine adverse effects} &
\begin{compactitem}
\item Le Covid a vraiment changé la donne. Personnellement, après avoir reçu le vaccin contre le Covid, mon immunité a diminué. Avant, je guérissais d'une grippe en une semaine, mais maintenant, cela prend au moins un mois.
\item Moi, mon ophtalmologue m'a critiqué car j'ai fait le vaccin contre le Covid. J'ai eu des problèmes dans mes yeux après avoir fait ce vaccin.
\item Le vaccin contre le COVID provoque Alzheimer.
\end{compactitem} &
\begin{compactitem}
\item Covid really changed things. Personally, after receiving the Covid vaccine, my immunity decreased. Before, I recovered from the flu in one week, but now it takes at least a month.
\item My ophthalmologist criticized me for taking the Covid vaccine. I had eye problems after receiving it.
\item The Covid vaccine causes Alzheimer’s.
\end{compactitem} \\

\midrule
\textbf{6 - Erosion of institutional trust post pandemic} &
\begin{compactitem}
\item Après la pandémie de COVID, l'opinion publique est devenue plus prudente.
\item Non, auparavant, nous avions une confiance aveugle dans le ministère de la Santé, mais les failles sont apparues après la pandémie de COVID, et notre confiance dans le ministère de la Santé a diminué...
\item Le COVID a créé ce problème, avant il n'y avait pas de précautions contre ceci ou cela. Depuis que nous avons appris les risques du vaccin et qu'il peut provoquer une crise cardiaque, les gens ont peur.
\end{compactitem} &
\begin{compactitem}
\item After the COVID pandemic, public opinion became more cautious.
\item No, previously we had blind trust in the Ministry of Health, but flaws became apparent after the COVID pandemic, and our trust in the Ministry of Health decreased...
\item COVID created this problem; before, there were no precautions about this or that. Since we learned about the risks of the vaccine and that it can cause a heart attack, people are afraid.
\end{compactitem} \\

\bottomrule
\end{tabular}
\end{scriptsize}
\vspace{12pt}
\caption{Representative utterances for topics with original French text and English translations.}
\label{tab:topics_appendix}
\end{table}
\end{landscape}

\bibliographystyle{plain}
\bibliography{references}

\end{document}